\documentclass[letterpaper, 10 pt, conference]{./ieeeconf}
\usepackage{times}
\usepackage[pdftex]{graphicx}
\usepackage{amsmath,amssymb,amsopn,amstext,amsfonts}
\usepackage{cancel}
\usepackage[space]{cite}
\usepackage{balance}
\usepackage{color}
\usepackage{mathtools}
\usepackage{algpseudocode}
\usepackage{algorithm}
\usepackage{bm}

\usepackage{diagbox}
\usepackage{float}
\usepackage{epstopdf}
\usepackage{pifont}
\usepackage{amsmath}
\usepackage{multirow}
\usepackage{url}
\usepackage{verbatim}
\usepackage{booktabs}
\usepackage{graphicx}
\usepackage{subcaption}
\usepackage{threeparttable}  
\usepackage{makecell}
\usepackage{soul}
\usepackage{xcolor}
\usepackage[normalem]{ulem}
\usepackage{etoolbox}
\usepackage[linkcolor=black,citecolor=black,urlcolor=black,colorlinks=true]{hyperref}

\graphicspath{{./figures/}}
\DeclareGraphicsExtensions{.png,.jpg,.eps,.pdf}
\IEEEoverridecommandlockouts
\newcommand{\delete}[1]{{\bgroup\markoverwith{\textcolor{red}{\rule[0.5ex]{2pt}{0.4pt}}}\ULon{#1}}}
\newcommand{\deletefig}[1]{{\bgroup\markoverwith{\textcolor{red}{\rule[2.5ex]{2pt}{2.0pt}}}\ULon{#1}}}

\setstcolor{red}
\hypersetup{draft}

\begin{document}
	\title{\LARGE \bf Autonomous and Adaptive Navigation for Terrestrial-Aerial \\ Bimodal Vehicles}
	\author{Ruibin Zhang\textsuperscript{1,2}, Yuze Wu\textsuperscript{1,2}, Lixian Zhang\textsuperscript{3}, Chao Xu\textsuperscript{1,2}, and Fei Gao\textsuperscript{1,2} 
	\thanks{This work was supported by National Natural Science Foundation of
		China under Grant 62003299 and Grant 62088101.}
	\thanks{\textsuperscript{1}State Key Laboratory of Industrial Control Technology, Zhejiang University, Hangzhou 310027, China.}
	\thanks{\textsuperscript{2} Huzhou Institute, Zhejiang University, Huzhou 313000, China.}
	\thanks{\textsuperscript{3} School of Astronautics, Harbin Institute of Technology, Harbin 150001, China.} 
	\thanks{E-mail:{\tt\small \{ruibin\_zhang, fgaoaa\}@zju.edu.cn}}}

	\maketitle
	\thispagestyle{empty}
	\pagestyle{empty}
	\begin{abstract}
	\label{sec:abstract}\textbf{}
	Terrestrial-aerial bimodal vehicles bloom in both academia and industry because they incorporate both the high mobility of aerial vehicles and the long endurance of ground vehicles. In this work, we present an autonomous and adaptive navigation framework to bring complete autonomy to this class of vehicles. The framework mainly includes 1) a hierarchical motion planner that generates safe and low-power terrestrial-aerial trajectories in unknown environments and 2) a unified motion controller which dynamically adjusts energy consumption in terrestrial locomotion. 
	Extensive real-world experiments and benchmark comparisons are conducted on a customized robot platform to validate the proposed framework's robustness and performance. During the tests, the robot safely traverses complex environments with terrestrial-aerial integrated mobility, and achieves $7\times$ energy savings in terrestrial locomotion. Finally, we will release our code and hardware configuration for the reference of the community\footnote{https://github.com/ZJU-FAST-Lab/Terrestrial-Aerial-Navigation}.

	\end{abstract} 

	\IEEEpeerreviewmaketitle 
	
	\section{Introduction}
    \label{sec:Introduction} 
    
	In recent years, unmanned aerial vehicles (UAVs) are involved in more and more applications, such as aerial photography, search-and-rescue, and delivery.
Among them, quadrotors are most widely used due to their simple structure, high mobility, and vertical takeoff and landing (VTOL) capability.
	Moreover, progresses in autonomous navigation enable quadrotors to fly safely and aggressively in unknown cluttered environments with full autonomy\cite{boyu2019ral, zhou2020ego, ye2020tgk}, greatly expanding their application area.
	
	However, quadrotor inherently suffers from sub-optimal power utilization (PU) because most of the energy is wasted on counteracting the body weight. 
This defect limits quadrotors' use in long-distance missions such as search-and-rescue, delivery, and active exploration. 	
	In such missions, the mobility of quadrotors is necessary for traversing extreme terrains, while the relatively short endurance can hardly support quadrotors to complete the entire mission. 
In contrast, although unmanned ground vehicles (UGVs) cannot cross rugged terrains, they enjoy a much better PU because the driving force only needs to overcome friction, not support their own weight. 
To combine the advantages of both types of mobile robots, researchers propose various terrestrial-aerial bimodal vehicles (TABVs) \cite{takahashi2015all, colmenares2019nonlinear, hada2017development ,nakao2019development,fan2019autonomous, li2021driving, Atay2021ControlAC, yamada2017development, kalantari2013design, kalantari2014modeling,kalantari2015hybrid,dudley2015micro, mizutani2015proposal, sabet2019rollocopter, tanaka2017design, choi2021baxter, tan2021multimodal, kalantari2020drivocopter, morton2017small, mintchev2018multi, david2021design}. TABV designs mostly consist of a quadrotor for aerial locomotion and additional mechanisms attached to it for terrestrial locomotion. The most common scheme, which is originally conceived by Kalantari et al.\cite{kalantari2013design, kalantari2014modeling, kalantari2015hybrid}, is to use passive mechanisms such as wheels\cite{takahashi2015all, hada2017development, colmenares2019nonlinear,fan2019autonomous,nakao2019development, li2021driving, Atay2021ControlAC}, cylindrical cages\cite{kalantari2013design, kalantari2014modeling, kalantari2015hybrid,  yamada2017development}, or spherical shells\cite{dudley2015micro, mizutani2015proposal, sabet2019rollocopter}. There are also other interesting designs that use motor-driven wheels \cite{tanaka2017design, choi2021baxter, kalantari2020drivocopter, tan2021multimodal} or deformable mechanisms \cite{morton2017small, mintchev2018multi, david2021design}.
The locomotion modes of TABVs can be dynamically adjusted depending on the need for better PU or higher mobility. 

	\begin{figure}
	\centering
	\includegraphics[width=1\linewidth]{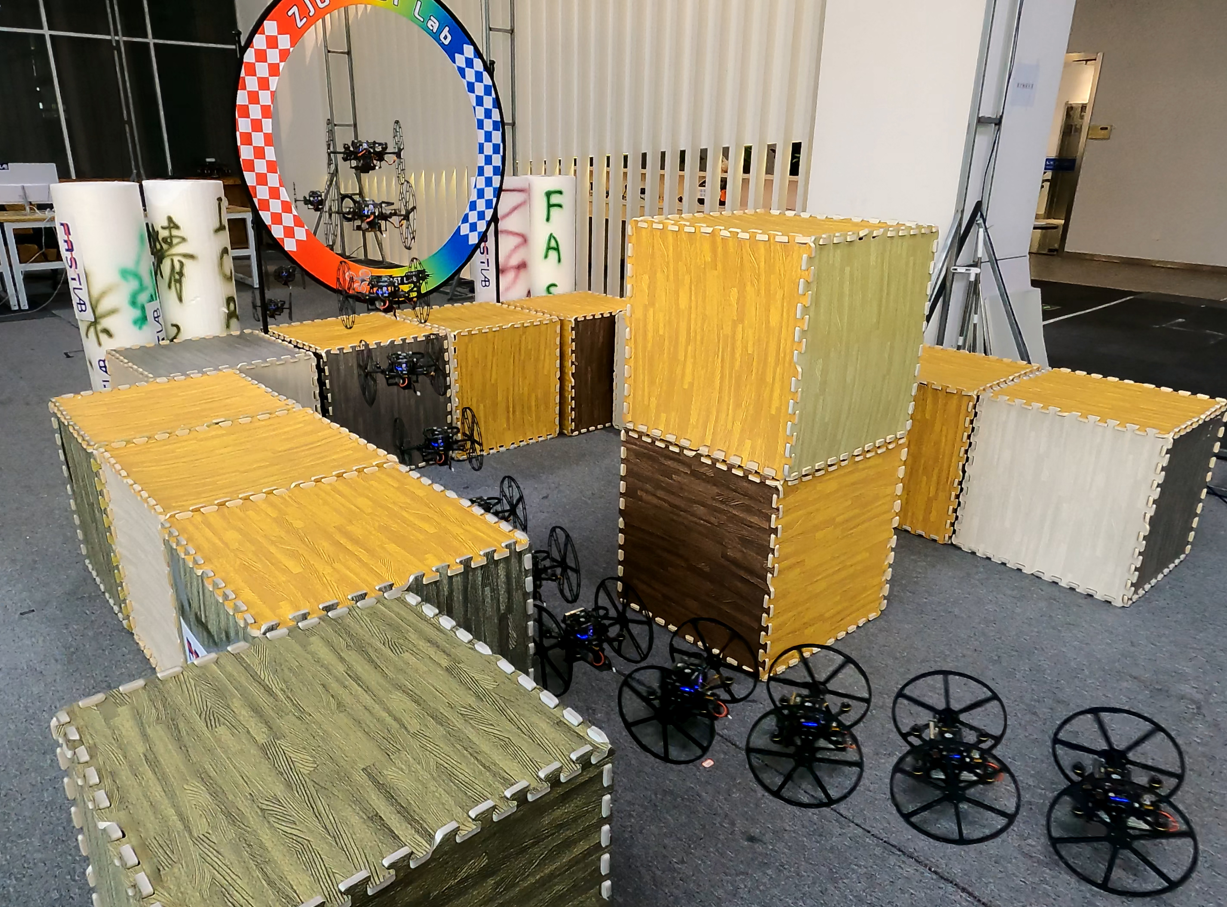}
	\captionsetup{font={small}}
	\caption{
			Demonstration of the proposed navigation framework, in which the customized TABV autonomously navigates the environment with terrestrial-aerial hybrid locomotion. Video is available at https://youtu.be/Bdb5mK9OKIo.
	}
	\label{pic:TIE_diagram}
	\vspace{-0.35cm}
\end{figure}
	

\begin{figure*}[t]
	\centering
	\includegraphics[width=1\linewidth]{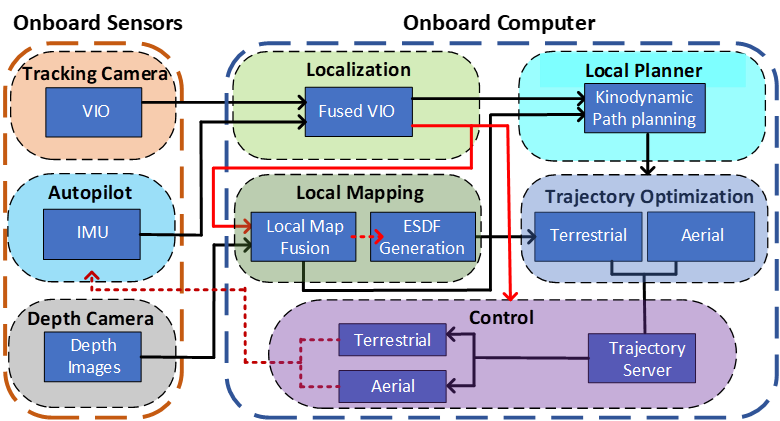}
	\captionsetup{font={small}}
	\caption{
		Software architecture. The perception, planning, and control modules run parallelly using onboard sensing and computing resourcess.
	}
	\label{pic:software_architecture}
	\vspace{-0.1cm}
\end{figure*}

 	In this work, the main focus is the autonomous navigation of TABVs, which is rarely covered by previous works. 	
	The proposed navigation framework endows TABVs with complete autonomy (as demonstrated in Fig.\ref{pic:TIE_diagram}). We adopt the passive-wheeled configuration \cite{kalantari2015hybrid} because it has the advantages of unified actuation system and simple structure. Firstly, we develop a hierarchical motion planner that searches for kinodynamic terrestrial-aerial hybrid paths and then refines them into safe, smooth, and dynamically feasible trajectories through B-spline optimization. The planning results are also energy-efficient because terrestrial trajectories are preferable unless the TABV has to fly over extreme terrains. Then, we design a unified terrestrial-aerial controller which includes an adaptive thrust adjustment method to improve PU in terrestrial locomotion. The results show up to \textbf{7 times less} energy consumption compared with aerial locomotion. 
	In addition, we develop self-localization and local map fusion modules that are applicable to terrestrial-aerial navigation scenarios. The overall software architectrue is shown in Fig.\ref{pic:software_architecture}. We also customize a TABV based on the design of Kalantari et al.\cite{kalantari2013design, kalantari2014modeling, kalantari2015hybrid} and equip it with adequate sensing and computing resources while ensuring portability and maneuverability.
	
	We perform sufficient experiments in challenging real-world environments to show the performance and robustness of the proposed navigation framework.
	During the tests, the TABV plans safe and low-power trajectories in unstructured dense environments and accurately tracks these trajectories even when there are sharp turnings. We also compare our work with cutting-edge works. The results show that the proposed methods are superior in planning performance, controlling accuracy, and PU.
	Contributions of this letter are:
	\begin{itemize}
	
	\item [1)] 
	A bi-level motion planner which generates safe, smooth, and dynamically feasible terrestrial-aerial hybrid trajectories.
	\item [2)]
	A unified motion controller for terrestrial-aerial locomotion, which dynamically adjusts the total thrust in terrestrial locomotion for better PU.
	
	\item [3)]
	 Combining the proposed methods with localization and mapping modules, releasing the source code and the customized robot platform as an open-source systematic solution.
	\end{itemize}

	\section{Related Work} 
	\label{sec:related_works}
	\subsection{Motion Planning for TABVs}
	\label{sec:related_hardware}	

	As stated before, only a few researchers have worked on autonomous navigation for TABVs. To the best of our knowledge, only Fan et al.\cite{fan2019autonomous} involve terrestrial-aerial motion planning. Firstly, it uses A* to search for a geometric path as guidance. By adding an extra energy cost to nodes in the air, this method tends to search for a terrestrial path. 
	Then, a waypoint is selected along this guiding path as the goal for a primitive-based local planner. It generates a set of minimum-snap trajectories, and scores each with a predefined cost to choose the best one. However, the path searching method is too coarse due to the lack of dynamic models. Also, since no post-refinement is applied to the trajectory in the local planner, its smoothness and dynamic feasibility cannot be guaranteed. Moreover, it does not consider the nonholonomic constraint in terrestrial locomotion. In the proposed planning method, we use kinodynamic path searching instead, and formulate a nonlinear optimization problem to refine the kinodynamic path. Apart from smoothness, collision avoidance, and dynamical feasibility cost, we also add a curvature limit cost for terrestrial trajectories in the optimization formulation to handle the nonholonomic constraint.

	\subsection{Motion Control for TABVs}
	 Several works \cite{ fan2019autonomous, colmenares2019nonlinear, takahashi2015all, Atay2021ControlAC} present control systems for passive-wheeled TABVs.
	Fan et al.\cite{fan2019autonomous} and Colmenares et al.\cite{colmenares2019nonlinear} propose cascaded control schemes similar to a general quadrotor controller.	 
	They both set the thrust as a constant value lower than the vehicle weight, so the PU cannot be improved dynamically. In fact, the total thrust can be flexibly adjusted because the ground support force partially shares the vehicle's weight.
	Takahashi et al.\cite{takahashi2015all} propose a controller based on Linear Quadratic Regulator (LQR) with online parameter estimation. Nevertheless, no real-world trajectory tracking experiments are presented to validate the method's efficacy.
	Atay et al.\cite{Atay2021ControlAC} extend the works of \cite{kalantari2013design, takahashi2015all} by elaborating on the specific dynamic model and developing a model-based control system. In addition, a thrust-optimization method is proposed. 
	However, this work takes the pitch angle as one of the flat outputs of the controller, but fails to present the mapping from a given trajectory to the flat outputs, making this control system inapplicable to trajectory tracking. The proposed control method extends the works of Fan et al.\cite{fan2019autonomous} and Colmenares et al.\cite{colmenares2019nonlinear} by presenting a thrust adjustment method. It calculates the total thrust according to the magnitude of desired control inputs so as to improve PU.

	 \begin{figure}[t]
	\centering
		\includegraphics[width=1\linewidth]{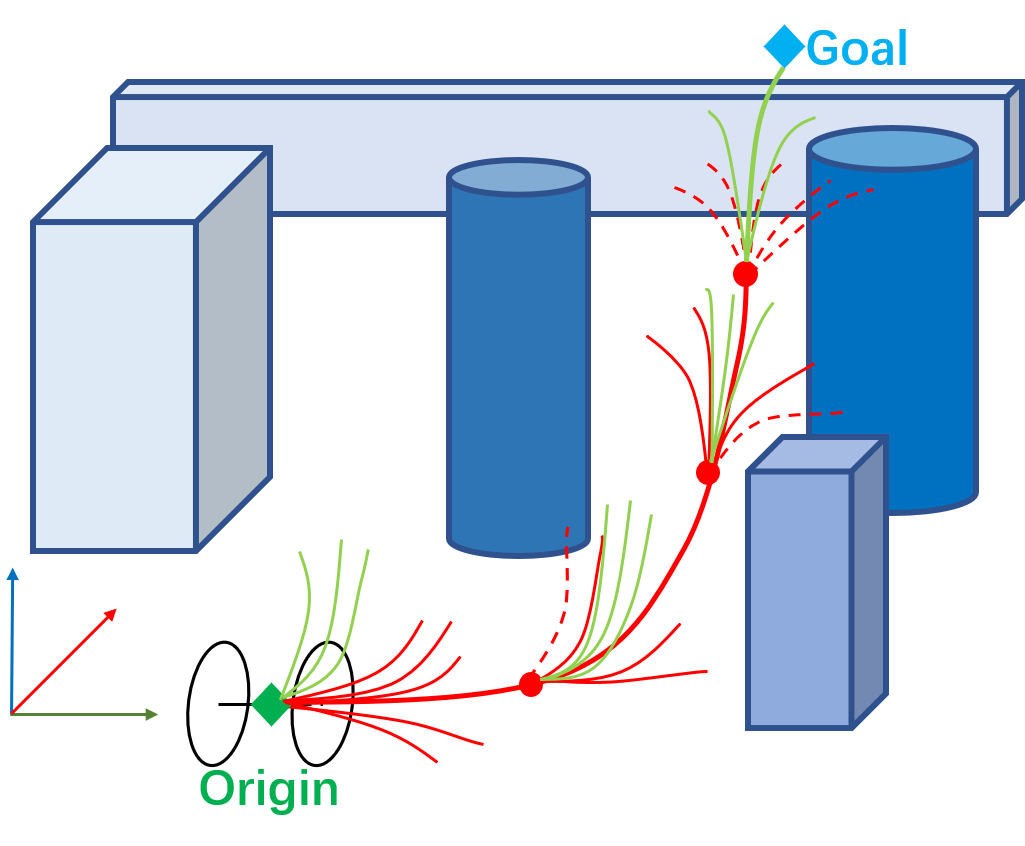}
	\captionsetup{font={small}}
	\caption{
		Illustration of the kinodynamic path searching method. Red curves represent the terrestrial motion primitives, while green curves are the aerial motion primitives. The method keeps planning aerial trajectories until an unavoidable obstacle appears.
	}
	\label{pic:astar}
	\vspace{0cm}
	\end{figure}

	\section{Safe Terrestrial-Aerial Motion Planning}
	The proposed terrestrial-aerial motion planner is built on Fast-Planner\cite{boyu2019ral}, which consists of a kinodynamic path searching method and a gradient-based spline optimizer. 
	The path searching method is based on hybrid-state A* algorithm\cite{dolgov2008practical}, which uses motion primitives instead of straight lines as graph edge in the searching loop. 	
	This work adds an extra energy consumption cost to the motion primitives whose destinations are above the ground.  Consequently, the path searching tends to plan terrestrial trajectories unless the TABV encounters enormous obstacles and needs to fly over them, as shown in Fig.\ref{pic:astar}.
	
	In trajectory optimization, we reparameterize the generated trajectory as a $p_{b}$ degree uniform B-spine with control points $\mathbf{Q}=\{\mathbf{Q}_{0}, \mathbf{Q}_{1}, ..., \mathbf{Q}_{N}\}$. 	Note that in terrestrial locomotion, we assume that the TABV moves on flat ground, so that the vertical motion can be omitted. We then classify the control points above the ground as $\mathbf{Q}_{a}$, and the rest as $\mathbf{Q}_{t} = \{\mathbf{Q}_{t0}, \mathbf{Q}_{t1},...,\mathbf{Q}_{tM}\}$. Each terrestrial control point is two-dimensional, i.e., $\mathbf{Q}_{ti} = (x_{ti}, y_{ti}), i \in [0, M]$.
	To refine the trajectory, we firstly adopt the following cost terms designed by Zhou et al.\cite{boyu2019ral}: 
	\begin{equation}
     f_{1} = \lambda_s f_s + \lambda_c f_c + \lambda_f (f_v + f_a), 
	\end{equation}
	where $f_s$ is the smoothness cost designed as an elastic band cost function. $f_c$ is the collision cost based on the ESDF gradient information. $f_v$ and $f_a$ are dynamical feasibility costs that limit velocity and acceleration. $\lambda_s, \lambda_c, \lambda_f$ are weights for each cost terms. 
	Due to the convex hull property of the B-spline, the above cost terms only constrain the control points $\mathbf{Q}$ for safety and dynamical feasibility.
	We refer the readers to \cite{boyu2019ral} for detailed formulations.
	
	    \begin{figure}[t]
		\centering
		\includegraphics[width=1\linewidth]{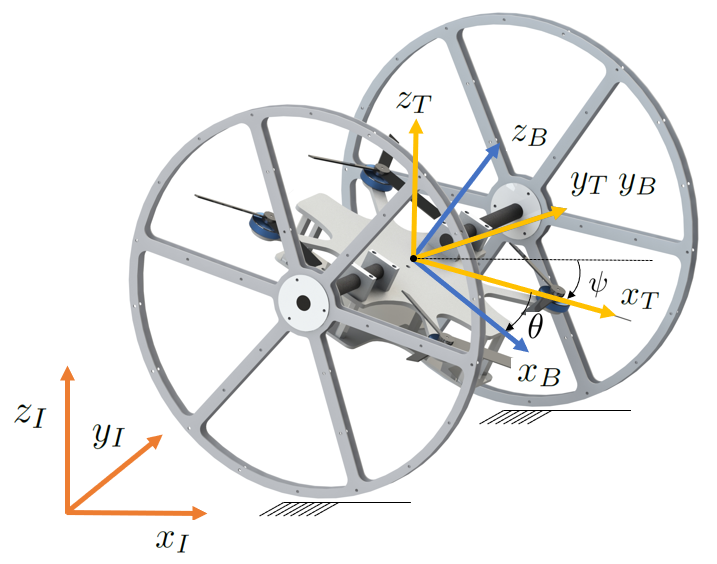}
		\captionsetup{font={small}}
		\caption{
			Diagram of the reference frames: inertial frame (I), body-fixed frame (B), and terrestrial frame (T). T is also a body-fixed frame with z-axis parallel to that of I, both pointing in the opposite direction of the gravity vector. Thus, T is separated from B by the rotation $\theta$ along y-axis.
		}
		\label{pic:frame}
		\vspace{0cm}
	\end{figure}

	In terrestrial locomotion, the TABV's velocity is limited to be parallel with the yaw angle due to the nonholonomic constraint. Therefore, if the trajectory is too curved, huge tracking errors will occur during turning.
	To resolve this, we enforce a cost on $\mathbf{Q}_{t}$ to limit the curvature of the terrestrial trajectory. 
	The curvature at $\mathbf{Q}_{ti}$ is defined as $\mathbf{C}_i = \frac{\Delta{\beta}_i}{\Delta{\mathbf{Q}_{ti}}}$, where 
	$\Delta{\beta}_i = |tan^{-1}\frac{\Delta{y_{ti+1}}}{\Delta{x_{ti+1}}} - tan^{-1}\frac{\Delta{y_{ti}}}{\Delta{x_{ti}}}|$, and $\Delta{\mathbf{Q}_{ti}} = \mathbf{Q}_{t} - \mathbf{Q}_{ti-1}$. 
	Therefore, this cost can be formulated as
	\begin{equation}
	f_{n} = \sum_{i=1}^{M-1}{F_n(\mathbf{Q}_{ti})},
	\end{equation}
	where $F_n(\mathbf{Q}_{ti})$ is a differentiable cost function with $\mathbf{C}_{max}$ specifying the curvature threshold:
	\begin{gather}
	F_n(\mathbf{Q}_{ti}) = \left\{
	\begin{aligned}
	& (\mathbf{C}_i - \mathbf{C}_{max})^2 								
	& \mathbf{C}_i > \mathbf{C}_{max},\\
	& 0 							
	& \mathbf{C}_i \leq \mathbf{C}_{max}.\\
	\end{aligned}
	\right.
	\end{gather}
	The derivation of the gradient can be found in \cite{dolgov2008practical}. Note that $\mathbf{Q}_{t}$ may be segmented into several subsets by intermediate aerial control points, the curvature of the endpoints are not taken into consideration. 
	In general, the overall objective function is formulated as follows:
	\begin{equation}
	f_{total} = \lambda_s f_s + \lambda_c f_c + \lambda_f (f_v + f_a) + \lambda_n f_n. 
	\end{equation}
	The optimization problem is solved by a non-linear optimization solver NLopt\footnote{https://nlopt.readthedocs.io/en/latest/}.
	
	After motion planning is done, a setpoint on the generated trajectory is selected according to the current timestamp, and then sent to the controller as a reference state in the inertial frame (defined in Fig.\ref{pic:frame}). 
	An aerial setpoint includes the yaw angle and 3D position, velocity, and acceleration ($[{^I}\mathbf{x}_a, {^I}\mathbf{v}_a, {^I}\mathbf{a}_a, {^I}\psi_a]^T$). A terrestrial one includes the yaw angle and 2D position and velocity ($[{^I}\mathbf{x}_t, {^I}\mathbf{v}_t, {^I}\psi_t]^T$). For consistency, ${^I}\psi_a$ and ${^I}\psi_t$ are both set to be parallel with the velocity.
	If the current setpoint is in a different locomotion mode than the previous one, an extra trigger will be sent to the controller for the locomotion switch. 
	
%

  	\section{Unified Terrestrial-Aerial Motion Control}
  	This section elaborates on the proposed controller, which adopts a cascaded architecture for both terrestrial and aerial locomotion. The reference frames are defined in Fig.\ref{pic:frame}. The estimated state obtained from onboard VIO is denoted as ${^I}\hat{\mathbf{X}}=[{^I}\hat{\mathbf{x}}, {^I}\hat{\mathbf{v}}, 
  	{^I}\hat{\Theta}, {^I}\hat{\dot{\Theta}}]^{T}$, including the position, velocity, orientation (parameterized by $Z-Y-X$ Euler angles ${^I}\hat{\Theta} = [{^I}\hat{\psi}, {^I}\hat{\theta}, {^I}\hat{\phi}]^T$) and its derivation (${^I}\hat{\dot{\Theta}} = [{^I}\hat{\dot{\psi}}, {^I}\hat{\dot{\theta}}, {^I}\hat{\dot{\phi}}]^T$). As for the locomotion switch, When take-off is desired, the controller immediately switches to aerial mode without a slow transition process. During landing, the controller commands the TABV to land smoothly with a constant speed, avoiding a sudden impact that may cause VIO divergence.

	  \begin{figure}[t]
	\centering
	\includegraphics[width=1\linewidth]{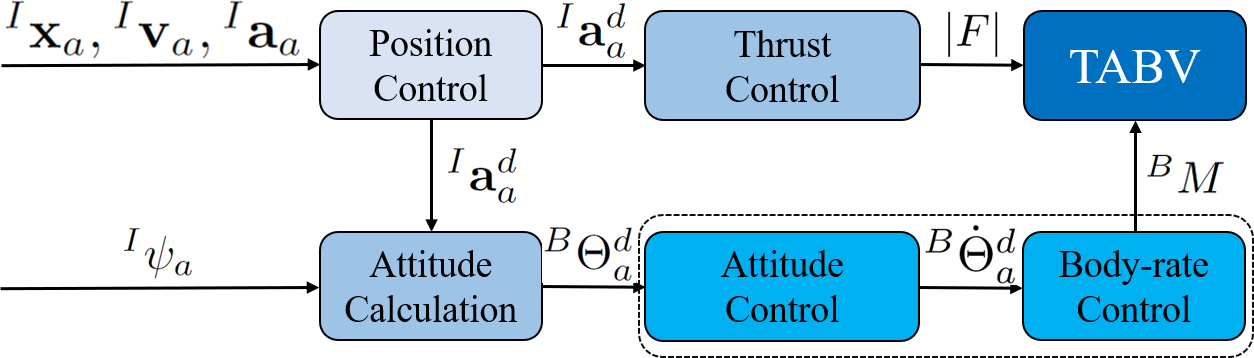}
	\captionsetup{font={small}}
	\caption{
		The cascaded controller for aerial locomotion.
	}
	\label{pic:Aerial_Controller}
	\vspace{0cm}
	\end{figure}

	  \subsection{Aerial Controller} 
	  The aerial controller is shown in Fig.\ref{pic:Aerial_Controller}. It takes the reference state $[{^I}\mathbf{x}_a, {^I}\mathbf{v}_a, {^I}\mathbf{a}_a, ^I\psi_a]^T$ from motion planning as the input. The position control module computes the position and velocity error using a proportional controller, and combines them with the reference term $^I\mathbf{a}_a$ to generate a desired acceleration $^I\mathbf{a}^d_a$. $^I\mathbf{a}^d_a$ is firstly used to generate the desired thrust $|F|$, and then used together with $^I\psi_a$ to calculate the desired attitude ${^B\Theta}^d_a$ leveraging the differential-flatness of quadrotors. The detailed equations of attitude calculation can be found in \cite{mellinger2011minimum}. The inner attitude control and body-rate control generate the attitude derivations $^B\dot{\Theta}^d_a$ and the desired moment $^B{M}$, respectively. These two modules are run on the onboard pilot using PX4 open-source firmware\footnote{https://github.com/PX4/PX4-Autopilot}. Finally, the PX4 firmware calculates the speed of each motor from $|F|$ and $^B{M}$, then send the signal to the actuators as the output.

	  \subsection{Terrestrial Controller} 
\begin{figure}[t]
	  	\centering
	  	\includegraphics[width=1\linewidth]{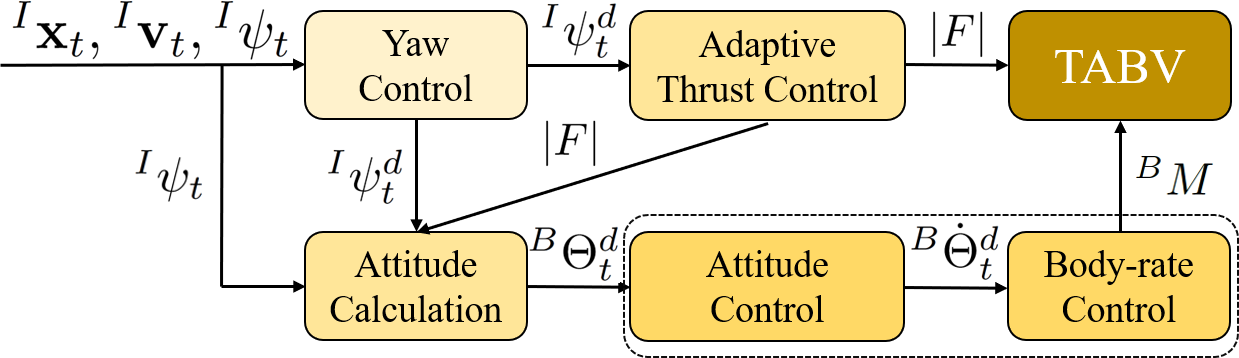}
	  	\captionsetup{font={small}}
	  	\caption{
	  		The cascaded controller for terrestrial locomotion.
	  	}
	  	\label{pic:Terrestrial_Controller}
	  	\vspace{0cm}
	  \end{figure}

\begin{figure}[t]
	\centering

	\includegraphics[width=1\linewidth]{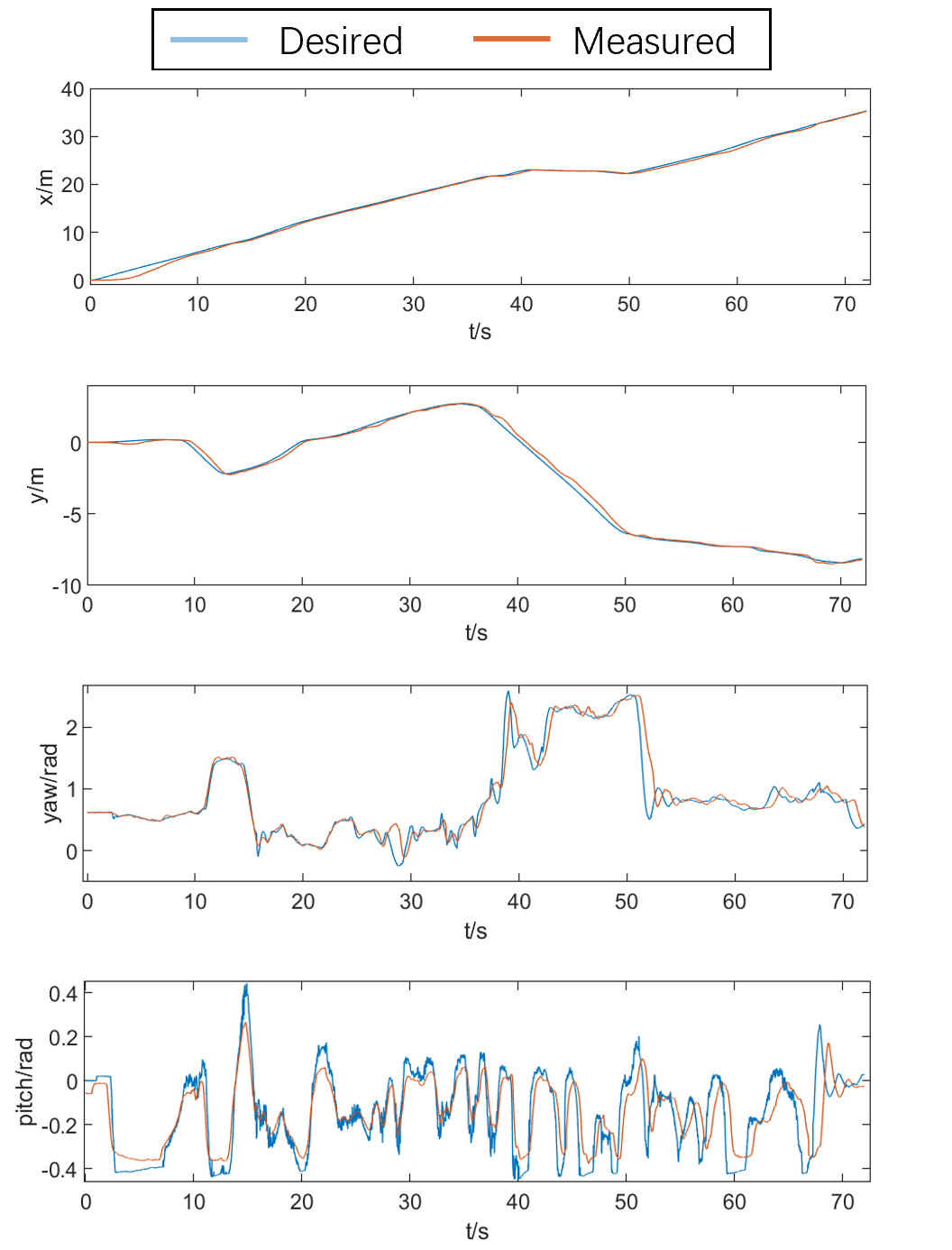}
	\captionsetup{font={small}}
	\caption{
		Tracking performance of the terrestrial controller in the office traversing experiment (Demonstrated in Sect.\ref{sec:experiments}). The results show that the planned trajectory and the desired attitude are closely tracked by the proposed terrestrial controller during the whole experiment. 
	}
	\label{pic:controller_performance}
	\vspace{-0.1cm}
	\end{figure}

	  Fig.\ref{pic:Terrestrial_Controller} illustrates the terrestrial controller, which owns a similar architecture to the aerial one. The attitude controller is executed by the onboard autopilot as well. The terrestrial controller's tracking performance is shown in Fig.\ref{pic:controller_performance}.
	  	 
	  \emph{\textbf{1) Yaw Control:}} 
	  The desired yaw $^I\psi^d_t$ is calculated according to the current position error between $^I\mathbf{x}_t$ and $^I\hat{\mathbf{x}}_t$, which is defined as $^I\mathbf{x}^e_t = {^I}\mathbf{x}_t - {^I}\hat{\mathbf{x}}_t $. When the norm of $^I\mathbf{x}^e_t$ is relatively small, $^I\psi^d_t$ is taken as the reference term $^I\psi_t$ which points along the trajectory's tangent direction. However, if the norm is larger than a threshold, $^I\psi^d_t$ is calculated to be parallel with $^I\mathbf{x}^e_t$ for error correction. The corresponding equations are shown as follows:
	  
	  \begin{gather}
	  ^I\psi^d_t = \left\{
	  \begin{aligned}
	  & ^I\psi_t							
	  & || {^I}\mathbf{x}^e_t || \leq ||\mathbf{x}^e||_m, \\
	  & \emph{tan}^{-1}\frac{{({^I}\mathbf{x}^e_t)_y}}{{({^I}\mathbf{x}^e_t)_x}} 							
	  & || {^I}\mathbf{x}^e_t || > ||\mathbf{x}^e||_m,\\
	  \end{aligned}
	  \right.
	  \end{gather}
	  
	  where $||\mathbf{x}^e||_m$ is the position error threshold. $({^I}\mathbf{x}^e_t)_x$ and $({^I}\mathbf{x}^e_t)_y$ are the x-axis and y-axis value of ${^I}\mathbf{x}^e_t$, respectively.
	  
	  \emph{\textbf{2) Adaptive Thrust Control:}} 
	  The desired total thrust $|F|$ is adaptively controlled according to the magnitude of current desired turning angle, defined as  $\Delta{\psi_t} = {^I}\psi^d_t - {^I}\hat{\psi}_t $. As mentioned before, position tracking error accumulates when the TABV is turning due to the nonholonomic constraint. To reduce the tracking error, we dynamically adjust the thrust so that it produces a maximal yaw acceleration ${\ddot{\psi}_{max}}$ large enough to make the TABV finish the turning in a short period $\Delta{t}$, which is set as $0.1s$ in experiments. Since $\Delta{t}$ is small, ${\ddot{\psi}_{max}}$ almost remains constant, and the yaw kinematics can be derived:
	  
	  \begin{align}	
	  \Delta{\psi_t} = {^I}\hat{\dot{\psi}}_t\Delta{t}+\frac{1}{2}\ddot{\psi}_{max}{\Delta{t}^2}.
	  \label{eq:des_yaw_acc} 
	  \end{align}
	  
	  Due to the nonlinearity of the inner attitude controller, 
	  we obtain the relationship between $|F|$ and ${\ddot{\psi}_{max}}$ by experimental fitting, given as
	  
	  \begin{equation}
	  |F|=0.04603\ddot{\psi}_{max}+0.0798,
	  \label{eq:F_yaw_relationship} 
	  \end{equation}
	  where $|F|$ is a normalized value between 0 and 1, and the dimension of ${\ddot{\psi}_{max}}$ is $rad/s^2$. $|F|$ can be solved from Equ.\ref{eq:des_yaw_acc} and Equ.\ref{eq:F_yaw_relationship}. Also, $|F|$ is ensured to be lower than hover thrust.

%
%
 
%

	  \emph{\textbf{3) Attitude Calculation:}} 
	  
	  It generates the desired attitude $^B\Theta^d_t = [{^B}{\psi}^d_t, {^B}{\theta}^d_t, {^B}{\phi}^d_t]^T$. We firstly calculate the desired attitude in inertial frame $^I\Theta^d_t = [{^I}{\psi}^d_t, {^I}{\theta}^d_t, {^I}{\phi}^d_t]^T$ and obtain $^B\Theta^d_t$ by coordinate transformation. Among them, ${^I}{\psi}^d_t$ has been computed in the yaw controller, and ${^I}{\phi}^d_t$ remains zero due to the assumption that the TABV moves on flat ground. The following equations give the derivation of ${^I}{\theta}^d_t$.
	  Firstly, we compute the x-axis value of the desired acceleration in terrestrial frame (denoted as $(^T\mathbf{a}^d_t)_x$) based on $^I\mathbf{x}^e_t$ and the velocity error $^I\mathbf{v}^e_t = {^I}\mathbf{v}_t - {^I}{\hat{\mathbf{v}}}_t$ with a feedback control law:
	  \begin{align}	
	  	(^T\mathbf{a}^d_t)_x = K_v(||{^I}\mathbf{v}^e_t|| + K_p||{^I}\mathbf{x}^e_t||)+K_I\delta,
	  \end{align}
	  where $\delta$ is the integral velocity tracking error. $K_V$, $K_P$ and $K_I$ are constant gains.
	  Then, ${^I}{\theta}^d_t$ can be calculated with the following dynamics equation. In this work, we do not take into account external forces such as the rolling friction.
	  \begin{align}
	  	{^I}{\theta}^d_t = sin^{-1}(\frac{M(^T\mathbf{a}^d_t)_x}{k_f|F|}),
	  \end{align}
	  where $M$ is the total mass of the TABV, and $k_f$ is the scale parameter.

 \section{System Integration} 
\label{sec:overview}
\subsection{Robot Platform} 
\label{sec:robot_platform}

\begin{figure}[t]
	\centering
	\includegraphics[width=1\linewidth]{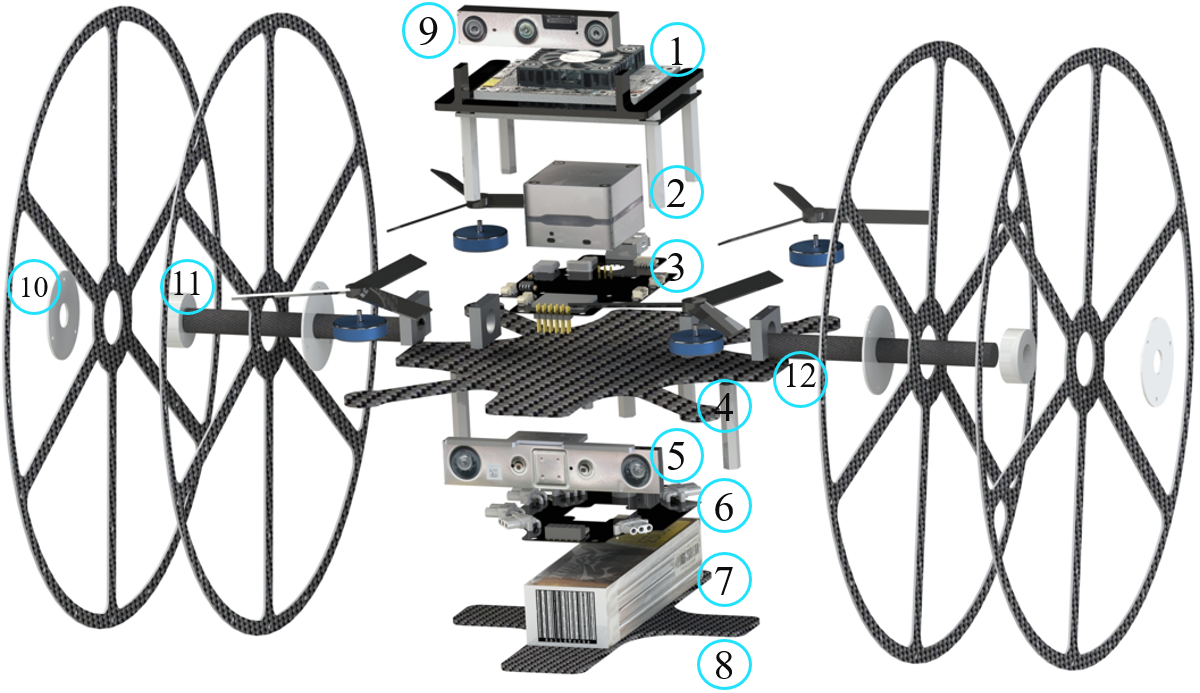}
	\captionsetup{font={small}}
	\caption{
		The detailed composition of the robot platform. The serial numbers represent (1) onboard computer, (2) autopilot, (3) upper PCB, (4) upper quadrotor frame, (5) tracking camera , (6) power-supply PCB, (7) battery, (8) lower quadrotor frame, (9) tracking camera, (10)  wheel, (11) bearing, (12) shaft.
		\vspace{0cm}
	}
	\label{pic:exploded_configuration}
\end{figure}

	We take a micro quadrotor with a diagonal wheelbase of 200mm as the main body of the TABV platform. For terrestrial locomotion, we connect each passive wheel to a shaft fixed on the quadrotor, so that each wheel can rotate freely relative to the quadrotor. For strength and weight considerations, we use carbon fiber as the main structure of the TABV, including the quadrotor frame, shafts, and wheels. The wheels, bearings, and shafts weigh 140g. The overall weight of the robot is 847.7g, including a 2300 mAh - 14.8 V battery that weighs 235g. The size of the robot is $280\times250\times250mm$. It can hover up to 9 minutes in aerial locomotion. The detailed composition of the robot platform is shown in the exploded diagram Fig.\ref{pic:exploded_configuration}.

	For autonomous navigation, we equip the TABV with the following onboard devices:

\begin{itemize}
	
	\item 
	
	RealSense D430 depth camera \footnote{https://store.intelrealsense.com/buy-intel-realsense-depth-module-d430-and-d4-board-bundle.html}: This camera provides the depth images for local map fusion.
	
	\item 
	
	RealSense T261 tracking camera\footnote{https://store.intelrealsense.com/buy-intel-realsense-tracking-module-t261.html}: This camera  provides robust Visual Inertial Odometry (VIO) for UAV state estimation.
	
	\item 
	
	CUAV V5+ autopilot:\footnote{http://doc.cuav.net/flight-controller/v5-autopilot/en/v5+.html}: It provides onboard IMU measurements and serves as the inner-loop controller.
	
	\item 
	
	Jetson Xavier NX\footnote{https://developer.nvidia.com/embedded/jetson-xavier-nx}: It is an onboard computer with 6-core NVIDIA Carmel CPU and 8GB RAM. The entire pipeline, including map fusion, state estimation, motion planning and control modules, runs on it.

\end{itemize}

\subsection{Software Architecture} 
\label{sec:architecture}

	The architecture of the proposed navigation framework is illustrated in Fig.\ref{pic:software_architecture}. Firstly, Visual Inertial Odometry (VIO) is obtained from the tracking camera. We fuse it with the IMU onboard the autopilot by Extended Kalman Filter (EKF) to generate smoother UAV state estimation. Then, the depth images from the depth camera are projected to the world frame as a point cloud. We then adopt a column-wise evaluation \cite{shan2018lego} to extract ground points. When maintaining an occupancy grid map, these points are not used, in order to avoid situations that the flat ground is set to be occupied. We also compute and update a Euclidean Signed Distance Field (ESDF) by an efficient algorithm developed by \cite{boyu2019ral}. 
	Afterward, the local planner searches for a kinodynamic path using the Fused VIO and the occupancy map. The resulted path is then optimized utilizing the gradient information obtained from the ESDF. The controller finally tracks the desired trajectory with both terrestrial and aerial locomotion. The experiments shown in Sect.\ref{sec:experiments} validate the real-time performance of the proposed navigation framework.


	\section{Results}

	\begin{figure}[t]
	\centering
	\includegraphics[width=1\linewidth]{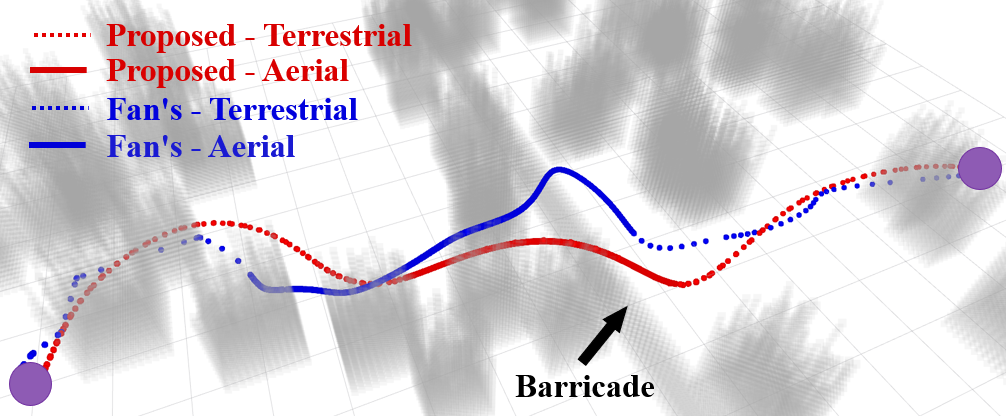}
	\captionsetup{font={small}}
	\caption{
		An instance of terrestrial-aerial trajectories generated by different methods. The 3D obstacles are set transparent to provide better views.
	}
	\label{pic:benchmark_diagram}
	\vspace{0.35cm}
\end{figure}	

	\begin{figure}[t]
		\centering
		\includegraphics[width=0.9\linewidth]{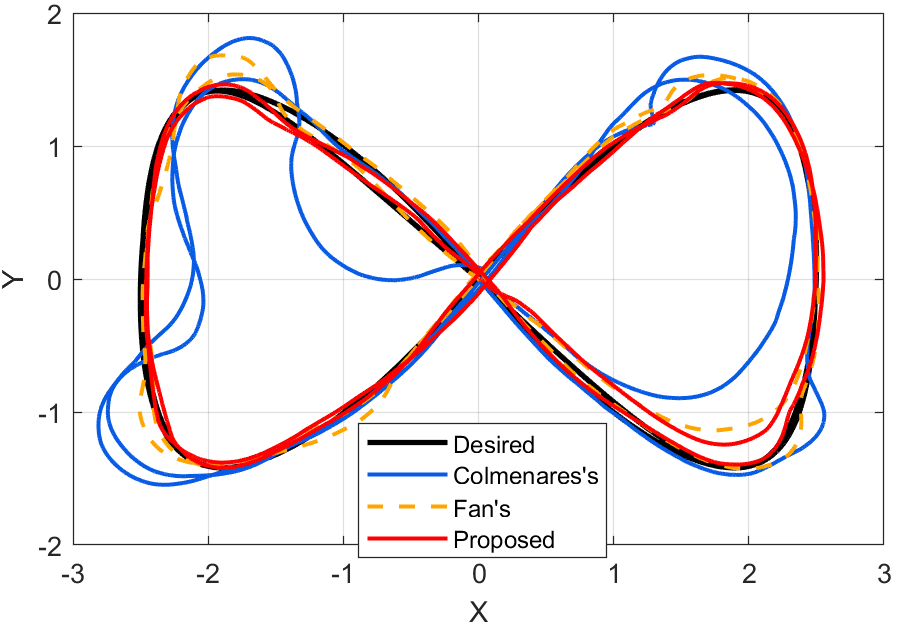}
		\captionsetup{font={small}}
		\caption{
			Comparison of different methods when tracking a lemniscate trajectory. This is the result when velocity limit is $1.2m/s$.
		}
		\label{pic:bazi}
		\vspace{-0.1cm}
	\end{figure}

	\subsection{Benchmark Comparisons} 
	\label{sec:benchmark}
	\begin{figure*}[t]
		\centering
		\includegraphics[width=1\linewidth]{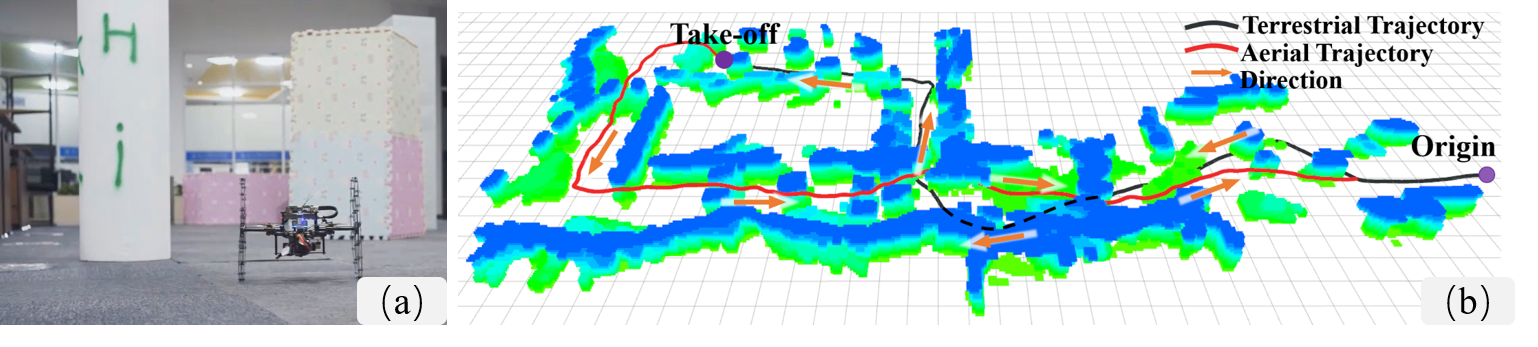}
		\captionsetup{font={small}}
		\caption{
			Experiment in a large office. The TABV moves about 100m with terrestrial-aerial locomotion, and the average velocity is about $0.8m/s$. 
		}
		\label{pic:robust_test_scene}
		\vspace{0.15cm}
	\end{figure*}	
	To demonstrate the superiority of the proposed navigation framework, we conduct benchmark comparisons against the previous works on terrestrial-aerial navigation in two-folds: the terrestrial-aerial planning and the terrestrial controller. 

	\emph{\textbf{1) Comparison of Terrestrial-Aerial Planning:}}
	We conduct comparisons between the proposed planning method and Fan's \cite{fan2019autonomous}.  Specifically, each algorithm runs for 50 times independently in a $20\times20\times3m$ simulation environment with 80 randomly deployed obstacles. We only compare planning methods and do not include terrestrial-aerial motion controllers in the simulation tests. The distance between the starting and goal positions is $16m$. We also set up a huge barricade between the starting and goal positions, requiring the robot to fly over it. All the computations are done on a 2.9 GHz processor with 16 GB RAM. The velocity and acceleration
	limits are set as $3m/s$ and $4m/s^2$. As shown in Tab.\ref{tab:planner_cmp}, the proposed planner finds trajectories with less computing time, better smoothness (integral of the squared acceleration), and higher success rate. Firstly, our planner both refines the trajectories' smoothness and dynamical feasibility, which are not considered in Fan's \cite{fan2019autonomous}. As shown in Fig.\ref{pic:benchmark_diagram}, the trajectory generated by Fan's \cite{fan2019autonomous} method is sub-optimal. In addition, the motion-primitive based method in Fan's \cite{fan2019autonomous} is incomplete, which may fail to generate feasible trajectories when facing complex environments, resulting in a low success rate even with a higher computing time. Therefore, our method performs better in both time efficiency and planning performance.

	\begin{table}[]
	\centering
	\caption{Terrestrial-Aerial Planning Comparison}
	\setlength{\tabcolsep}{2.5mm}	
	
	\begin{tabular}{|c|c|l|c|c|c|c|}
		\hline
		\multirow{2}{*}{Method} & \multicolumn{2}{c|}{\multirow{2}{*}{\begin{tabular}[c]{@{}c@{}}Comp. \\ Time(ms)\end{tabular}}} & \multicolumn{3}{c|}{Inte. of Acc. $(m^2/s^3)$} & \multirow{2}{*}{\begin{tabular}[c]{@{}c@{}}Success \\ Rate(\%)\end{tabular}} \\ \cline{4-6}
		& \multicolumn{2}{c|}{}                                                                           & Mean         & Max         & Std        &                                                                              \\ \hline
		Proposed     & \multicolumn{2}{c|}{\textbf{3.36}}                                                                      & \textbf{73.01}        & \textbf{130.48}      & \textbf{20.50}      & \textbf{98}                                                                           \\ \hline
		Fan's \cite{fan2019autonomous}          & \multicolumn{2}{c|}{3.66}                                                                      & 250.10       & 367.86      & 46.69      & 74                                                                        \\ \hline
	\end{tabular}
	\vspace{0.25cm}	
	\label{tab:planner_cmp}	
\end{table}   

\begin{table}[t]	
	\centering	
	\caption{Terrestrial Controller Comparison}	
	\setlength{\tabcolsep}{2mm}	
	\renewcommand\arraystretch{1.2}	
	{		
		\begin{tabular}{|c|c|c|c|c|c|}			
			\hline			
			Velocity & Method & $T_n$ & $E_a$(m) & $E_m$(m) \\ \hline			
			& Colmenares's \cite{colmenares2019nonlinear} & 0.147 & 0.0404 & 0.3736 \\ \cline{2-5}			
			$0.8m/s$ & Fan's \cite{fan2019autonomous}       & 0.147 & 0.0600 & 0.4608 \\ \cline{2-5}			
			& Proposed     & 0.147(Average) & \textbf{0.0158} & \textbf{0.0946} \\ \hline			
			& Colmenares's \cite{colmenares2019nonlinear}& 0.153 & 0.0973 & 0.6690 \\ \cline{2-5}			
			$1m/s$   & Fan's \cite{fan2019autonomous}       & 0.153 & 0.0358 & 0.2725 \\ \cline{2-5}			
			& Proposed     & 0.153(Average) & \textbf{0.0225} & \textbf{0.1015} \\ \hline			
			& Colmenares's \cite{colmenares2019nonlinear}& 0.171 & 0.1339 & 0.5293 \\ \cline{2-5}			
			$1.2m/s$ & Fan's \cite{fan2019autonomous}       & 0.171 & 0.0474 & 0.3005 \\ \cline{2-5}			
			& Proposed     & {0.171(Average)} & \textbf{0.0338} & \textbf{0.1894} \\ \hline			
	\end{tabular}}	
	\label{tab:controller_cmp}	
	\vspace{-0.15cm}	
\end{table}
	
	\emph{\textbf{2) Comparison of Terrestrial Controller:}}
	We compare the proposed terrestrial controller with method\cite{colmenares2019nonlinear, fan2019autonomous} in real-world environments. Only the outer translation control is compared because it is our focus, and the inner attitude control of all methods is executed by the autopilot. During the comparison, the TABV uses each controller to track a lemniscate trajectory with different velocities. Since method\cite{colmenares2019nonlinear, fan2019autonomous} set the desired thrust to be constant, we first test our method, then calculate the desired average normalized thrust (denoted as $T_n$) and assign it to method\cite{colmenares2019nonlinear, fan2019autonomous}. The average and maximal trajectory tracking error $E_a$ and $E_m$ are compared. As shown in Tab.\ref{tab:controller_cmp} and Fig.\ref{pic:bazi}, the proposed method achieves lower $E_a$ and $E_m$ with the same $T_n$ in every case. With the adaptive thrust control, the proposed controller generates a larger thrust to make the TABV pass through sharp turnings faster, thereby reducing the tracking error. 
	In contrast, method \cite{colmenares2019nonlinear, fan2019autonomous} requires a larger thrust at all times to accurately track terrestrial trajectories, which greatly increases energy consumption. Fan et al.\cite{fan2019autonomous} state in the paper that the energy consumption in terrestrial locomotion is two-sevenths of the aerial locomotion during trajectory tracking, while the energy consumption ratio with the proposed controller is half of that (as shown in Sect.\ref{sec:experiments}). In the trajectory tracking test conducted by Colmenare et al.\cite{colmenares2019nonlinear}, the thrust is kept constant at $80\%$ of the robot's gravity, whereas the average thrust with the proposed controller is no more than $30\%$ of that. We experimentally test that the former consumes four times as much energy as the latter.

	\subsection{Experiments} 
	\label{sec:experiments}
\begin{figure}[t]
		\centering
		\includegraphics[width=1\linewidth]{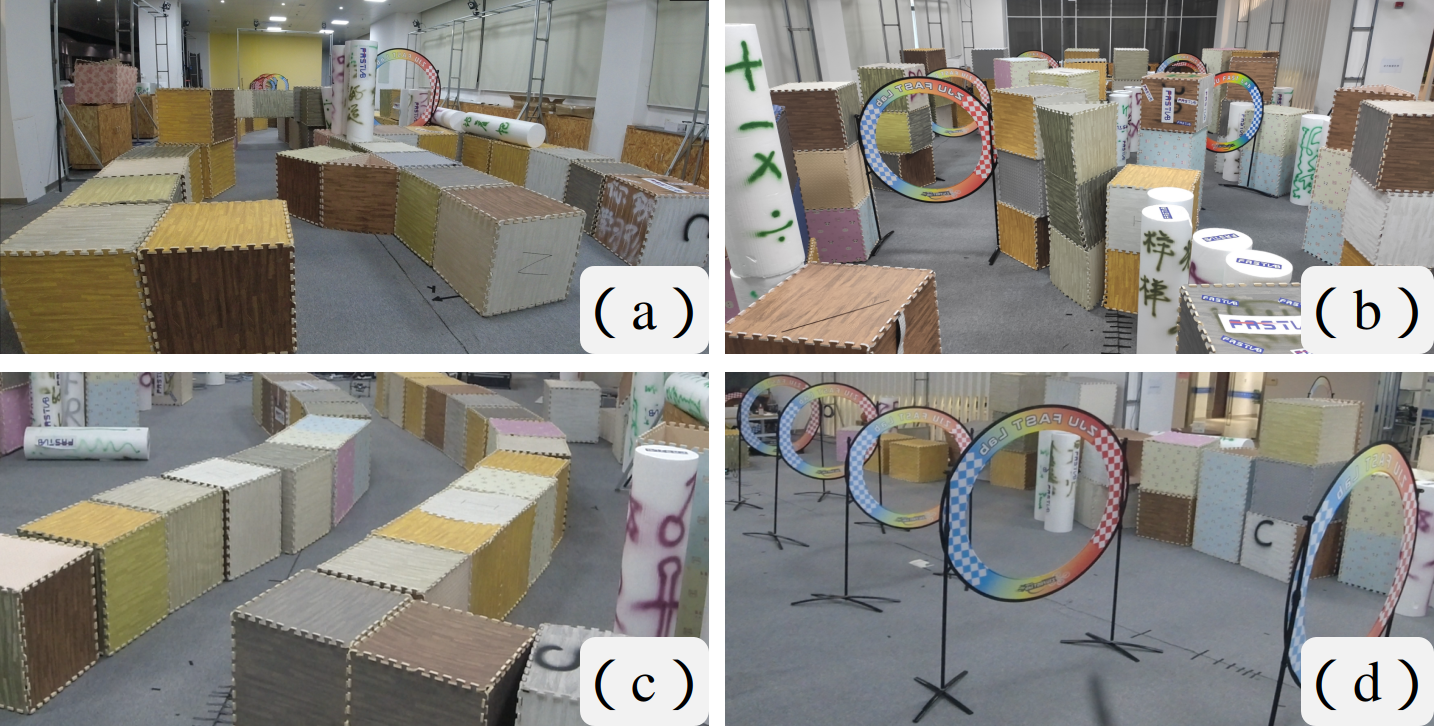}
		\captionsetup{font={small}}
		\caption{
			The real-world experimental scenes. a) The complex maze. b) The scene for comparison between terrestrial and aerial locomotion. c) Terrestrial winding tunnel. d) Aerial winding tunnel. These experiments are all carried out in a room filled with artificially placed cylindrical or circular obstacles.
		}
		\label{pic:experimental_scene}
		\vspace{-0.2cm}
	\end{figure}

	To demonstrate the autonomy and performance of the entire robot system, we perform extensive autonomous tests in various complex environments (as shown in Fig.\ref{pic:robust_test_scene} and Fig.\ref{pic:experimental_scene}). 
	Except for several waypoints, no prior information of the environments is given. The unknown dense environments and limited onboard vision make the experiments challenging. More details are included in the video.


	\emph{\textbf{1) Walking out of a Complex Maze:}}
	\begin{figure}[t]
		\centering
		\includegraphics[width=1\linewidth]{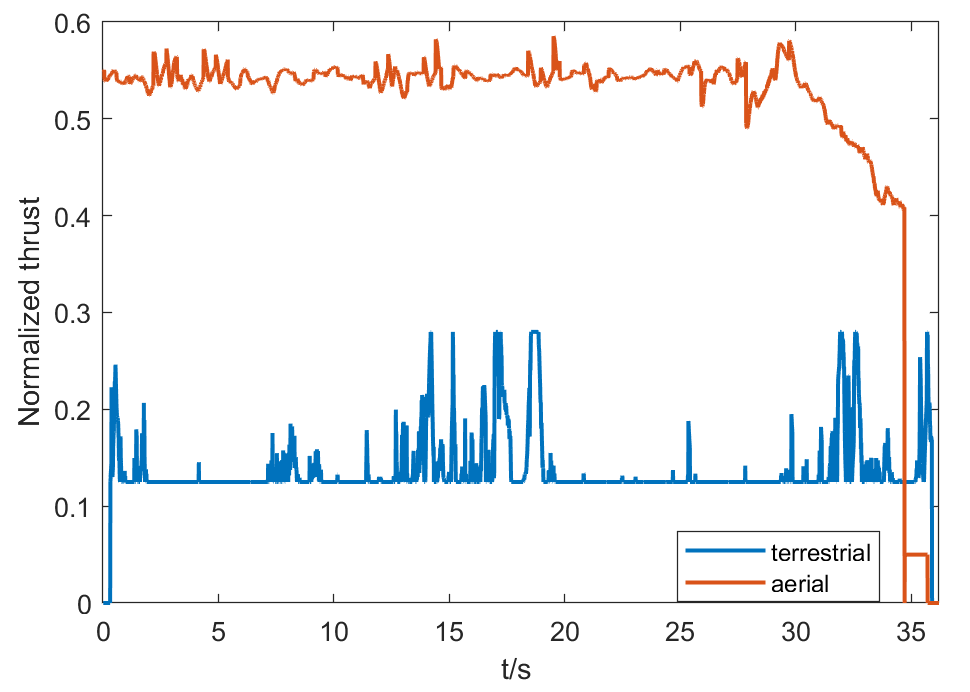}
		\captionsetup{font={small}}
		\caption{
			The thrust curve of both terrestrial and aerial locomotion. Note that the sharp drop in the air thrust curve is due to landing.
		}
		\label{pic:thrust_compare}
		\vspace{-0.3cm}
	\end{figure}
	In this experiment, the TABV has to navigate a complex maze with high obstacles (about $60cm$ in height), sharp turnings (the largest of which reaches $180^{\circ}$), and an unavoidable barricade. The velocity limit is set to be $0.8m/s$. It turns out that the TABV manages to walk out of this maze, and it remains in terrestrial locomotion except when it flies over the unavoidable barricade. This result is as expected because terrestrial locomotion is preferable due to better PU. 
	
	\emph{\textbf{2) Flying vs Rolling:}}	
	This experiment presents quantitative PU comparisons between terrestrial and aerial locomotion. The experimental scene is complicated as well because of tortuous paths and a great many obstacles. However, it is set to be passable for both terrestrial and aerial locomotion. In the experiment, the TABV passes through the environment in terrestrial and aerial locomotion with the same velocity limit $1m/s$, respectively. The normalized thrust curve is depicted in Fig.\ref{pic:thrust_compare}. The average thrust is $0.138$ in terrestrial locomotion and $0.522$ in aerial locomotion. That is, the TABV passes this challenging test in both locomotion modes, but requires about only a quarter as much thrust in terrestrial locomotion as in the aerial one. We also experimentally measure that the corresponding energy consumption ratio is approximately $\mathbf{1 : 7}$. This result highlights the great advantage of terrestrial locomotion in PU.
	
	\emph{\textbf{3) Moving through Winding Tunnels:}}
	We perform aggressive flight and rolling tests in this experiment to present the proposed system's high mobility even in autonomous navigation. Two winding tunnels are set for the flight and rolling test, respectively. The end of each tunnel is outside the TABV's sensing range, so it needs to replan in time and turn quickly to pass the test. As a result, the TABV can travel back and forth through the tunnels with a velocity up to $2m/s$ in aerial locomotion and $1.8m/s$ in terrestrial locomotion. The results are comparable with state-of-the-art autonomous quadrotor systems in \cite{boyu2019ral, zhou2020ego, ye2020tgk}.

	\emph{\textbf{4) Traversing a large Office:}}
	The last experiment is conducted in an unknown office with a size over $40\times20m$. It is full of cluttered objects, leaving only narrow passages, which brings difficulties to motion planning. What is more, the lighting and terrain condition around the office does not remain the same, posing a huge challenge to the perception and the terrestrial controller of the system. In order to test both terrestrial and aerial navigation, We set up a take-off waypoint halfway and keeps the TABV flying after it passes this waypoint. It turns out that the TABV safely traverses the office in terrestrial-aerial integrated locomotion. The executed trajectories are shown in  Fig.\ref{pic:robust_test_scene}. This experiment strongly demonstrates the robustness of the proposed system.
	
	
%
%
%
%
%
%
%
%

	\section{Conclusion}
	\label{sec:conclusion}
	TABVs possesses distinct advantages because they combine both the mobility of UAVs and the long endurance of UGVs. In this work, 
	we present a navigation framework that enables TABVs to safely navigate in unknown cluttered environments with terrestrial-aerial hybrid locomotion. 
	The framework mainly consists of a safe motion planner and a unified motion controller. We also incorporate localization and mapping modules and integrate the navigation framework into an onboard system on a customized terrestrial-aerial vehicle.
	We carry out challenging tests in multiple unknown dense environments to show the proposed system's performance and robustness. 
	
	For future work, we will pay attention to motion planning on uneven terrains. Besides, we will consider the exogenous forces in the controller so as to further improve the control accuracy and energy savings. Furthermore, we will develop active exploration algorithms leveraging the bimodal locomotion of TABVs \cite{Williams2020exploration}.

	\bibliography{RAL2022_zrb}
\end{document}